\documentclass[runningheads]{llncs}
\usepackage{cite}
\usepackage{amsmath,amssymb,amsfonts}
\usepackage{mathtools}
\usepackage{epsfig}
\usepackage[mathscr]{euscript}
\usepackage{color}
\usepackage{booktabs} 
\usepackage{stmaryrd} 
\usepackage{bbm} 
\usepackage{blindtext, comment}
\usepackage{rotating} 
\usepackage{tikz}
\usepackage{standalone}
\usetikzlibrary{arrows, patterns}
\usetikzlibrary{decorations.pathreplacing,angles,quotes}
\usepackage{framed, standalone}
\usepackage{graphicx}
\usepackage{capt-of}
\usepackage{balance}
\usepackage{makecell} 
\usepackage{afterpage}
\usepackage[linesnumbered,lined,boxed,commentsnumbered,ruled,vlined]{algorithm2e}
\usepackage{setspace}
\usepackage{textcomp}
\usepackage{xcolor}
\def\BibTeX{{\rm B\kern-.05em{\sc i\kern-.025em b}\kern-.08em
    T\kern-.1667em\lower.7ex\hbox{E}\kern-.125emX}}
\begin{document}
\title{VecHGrad for Solving Accurately \\Tensor Decomposition
}
%
\author{Jeremy Charlier \inst{1}
\and
Vladimir Makarenkov \inst{2}
}

\institute{}
\authorrunning{J. Charlier et al.}
\institute{National Bank of Canada, Montreal, Canada  \\
\email{\{name.surname@\}@bnc.ca}
\and
Université du Québec à Montréal (UQAM), Montreal, Canada  \\
\email{\{surname.name@\}@uqam.ca}
}

\maketitle

\begin{abstract}
Tensor decomposition is a collection of factorization techniques for multidimensional arrays. Today's data sets, because of their size, require tensor decomposition involving factorization with multiple matrices and diagonal tensors such as DEDICOM or PARATUCK2. Traditional tensor resolution algorithms such as Stochastic Gradient Descent (SGD) or Non-linear Conjugate Gradient descent (NCG), cannot be easily applied to these types of tensor decomposition or often lead to poor accuracy at convergence. We propose a new resolution algorithm, VecHGrad, for accurate and efficient stochastic resolution over all existing tensor decomposition. VecHGrad relies on the gradient, an Hessian-vector product, and an adaptive line search, to ensure the convergence during optimization. Our experiments on five popular data sets with the state-of-the-art deep learning gradient optimizers show that VecHGrad is capable of converging considerably faster because of its superior convergence rate per step. VecHGrad targets as well deep learning optimizer algorithms. The experiments are performed for various tensor decomposition, including CP, DEDICOM, and PARATUCK2. Although it involves an Hessian-vector update rule, VecHGrad's runtime is similar in practice to that of gradient methods such as SGD, Adam, or RMSProp.
\end{abstract}
\section{Motivation} \label{sec::intro}
Tensors are multidimensional, or N-order, arrays. Tensors are able to scale down a large amount of data to an interpretable size using different decomposition types, also called factorizations. The data sets used in machine learning \cite{globerson2007euclidean} and data mining \cite{zhouonline} are multi-dimensional and, therefore, tensors and their decompositions are highly appropriate \cite{kolda2009tensor}. Because of the size of modern data sets, Tensor Decompositions (TDs), such as the CP/PARAFAC \cite{carroll1970analysis,harshman1970foundations} decomposition, later denoted CP, are now challenged by other TDs such as DEDICOM \cite{bader2007temporal} and PARATUCK2 \cite{harshman1996uniqueness,charlier2018user}. CP decomposes the original tensor as a sum of rank-one tensors, as illustrated in figure \ref{fig::cptd}, whereas DEDICOM and PARATUCK2 decompose the original tensor as a product of matrices and diagonal tensors, as shown in figures \ref{fig::dedicomtd} and  \ref{fig::paratuck2td}, respectively. Depending on the TD type, different latent variables are highlighted with their respective asymmetric relationships. Fast and accurate tensor resolutions have however required specific numerical optimization methods known as preconditioning methods.  \\

Well-known in the Machine Learning (ML) and Deep Learning (DL) community, the standard Stochastic Gradient (SGD) optimization method \cite{robbins1951textordfemininea,kiefer1952stochastic} is widely used in different ML and DL approaches. It is however losing its momentum over recent preconditioning gradient methods. The latter considers a matrix, a preconditioner, to update the gradient before it is used. Standard preconditioning methods include Newton's method, which employs the exact Hessian matrix, and the quasi-Newton methods, which do not require the exact Hessian matrix \cite{wright1999numerical}. The computational cost of the exact Hessian matrix is one of the major limitations of Newton's method. 
Introduced to answer some of the challenges in ML and DL, AdaGrad \cite{duchi2011adaptive} uses as a preconditioner the co-variance matrix of the accumulated gradients. Due to the dimensions of the ML problems, 
specialized variants were proposed to replace the full preconditioning methods by diagonal approximation methods, such as Adam \cite{kingma2014adam}, 
or by other schemes, such as Nesterov Accelerated Gradient (NAG) \cite{nesterov2007gradient} or SAGA \cite{defazio2014saga}. The diagonal approximation methods are often preferred because of the super-linear memory consumption of other methods \cite{gupta2018shampoo}.  \\

In this work, 
we aim at efficient and accurate resolution of high-order TDs for which most of the ML and DL state-of-the-art optimizers fail. We describe how to exploit the quasi-Newton convergence with a diagonal approximation of the Hessian matrix and an adaptive line search. Our algorithm, VecHGrad for Vector Hessian Gradient, returns the tensor structure of the gradient using a separate preconditioner vector. 
Our analysis relies on the extensions of vector analysis to the tensor world. We show the superior capabilities of VecHGrad for various resolution algorithms, such as Alternating Least Squares (ALS) or Non-linear Conjugate Gradient (NCG) \cite{acar2011scalable}, and some popular DL optimizers, such as Adam or RMSProp. Our main contributions are summarized below: 
\begin{itemize}
    \item We propose a new resolution algorithm, VecHGrad, that uses the gradient, the Hessian-vector product and an adaptive line search to perform accurate and fast optimization of the objective function of TD.
    \item We demonstrate the superior accuracy of VecHGrad at convergence. We compare it, on five popular data sets, to other resolution algorithms, including deep learning optimizers, for the three most common TDs known as CP, DEDICOM, and PARATUCK2. 
\end{itemize}

The paper is structured as follows. We discuss the related work in Section \ref{sec::relwork}. We review briefly tensor operations and tensor definitions in Section \ref{sec::background}. 
We then describe in Section \ref{sec::propmethod} how VecHGrad performs in a numerical optimization scheme applied to tensor structures without the requirement of knowing the Hessian matrix. We highlight the experimental results in Section \ref{sec::exp}. Finally, we conclude and address promising directions for future work.

\section{Related Work} \label{sec::relwork}
VecHGrad uses a diagonal approximation of the Hessian matrix and, therefore, is related to other diagonal approximations such as AdaGrad \cite{duchi2011adaptive}, which is very popular and frequently applied \cite{gupta2018shampoo}. 
However, it only uses gradient information, as opposed to VechGrad which uses both gradient and Hessian information. Other approaches extremely popular in ML and DL include Adam \cite{kingma2014adam}, NAG \cite{nesterov2007gradient}, SAGA \cite{defazio2014saga}, and RMSProp \cite{hinton2012rmsprop}. The non-exhaustive list of ML optimizers is considered in our study case since it offers a strong baseline comparison for VecHGrad.  \\

Since our study case is related to TDs, the methods specifically designed for TDs have to be mentioned. The most popular optimization scheme among the resolution of TDs is the Alternating Least Squares (ALS). Under the ALS scheme \cite{kolda2009tensor}, one element of the decomposition is fixed. The fixed element is updated using the other elements. Therefore, all the elements are successively updated at each step of the iteration process until a convergence criteria is reached, e.g. a fixed number of iteration. 
For every TDs, there exists at least one ALS resolution scheme. The ALS resolution scheme was introduced in \cite{carroll1970analysis} and \cite{harshman1970foundations} for the CP decomposition, in \cite{harshman1978models} for the DEDICOM decomposition and in \cite{harshman1996uniqueness} for the PARATUCK2 decomposition. An updated ALS scheme was presented in \cite{bro1998multi} to solve PARATUCK2. In \cite{bader2007temporal}, Bader et al. proposed ASALSAN to solve with non-negativity constraints the DEDICOM decomposition with the ALS scheme. While some matrix updates are not guaranteed to decrease the loss function, the scheme leads to overall convergence. In \cite{charlier2018non}, Charlier et al. have recently proposed a non-negative scheme for the PARATUCK2 decomposition.  \\

Other approaches are specifically designed for one TD using gradient information. 
In \cite{acar2011scalable,acar2011all}, an optimized version of NCG for CP is presented, i.e. CP-OPT. In \cite{maehara2016expected}, an extension of the Stochastic Gradient Descent (SGD) is described to obtain, as mentioned by the authors, an expected CP TD. 
The performance on other TDs was, however, not assessed. The comparison to other numerical optimizers in the experiments was rather limited, especially when considering existing popular machine learning and deep learning optimizers. In contrast, VecHGrad is detached from any particular model structure, including the choice of TDs. It only relies on the gradient, the Hessian diagonal approximation and an adaptive line search, crucial for fast convergence of complex numerical optimization. Consequently, VecHGrad is easy to implement and use in practice as it does not require to be optimized for a particular model structure.  \\

\section{Background} \label{sec::background}
In this section, we briefly introduce the preliminaries of TDs. Scalars are denoted by lower case letters \textit{a}, vectors by boldface lowercase letters \textbf{a}, matrices by boldface capital letters \textbf{A}, and N-order tensors by the Euler script notation $\mathscr{X}$.

\subsection{Tensor Operations}
\textbf{Vectorization.} The vectorization operator flattens a tensor of $n$ entries to a column vector $\mathbb{R}^n$. The ordering of the tensor elements is not important as long as it is consistent \cite{kolda2009tensor}. For a third order tensor $\mathscr{X}\in\mathbb{R}^{I\times J \times K}$, the vectorization of $\mathscr{X}$ is equal to
$
  \text{vec} (\mathscr{X}) = 
  \begin{bmatrix} 
    x_{111} & x_{112} & \cdots & x_{IJK}
  \end{bmatrix}^T \;.
$

\textbf{Tensor Norm.} The square root of the sum of all squared tensor entries of the tensor $\mathscr{X}$ defines its norm:
$
\parallel \mathscr{X}\parallel \:=\: \sqrt{\sum_{j=1}^{I_1}\sum_{j=2}^{I_2} ... \sum_{j=n}^{I_n}x_{j_1, j_2, ..., j_n}^2} \;.
$

\textbf{Tensor Rank.} The rank-\textit{R} of a tensor $\mathscr{X}\in\mathbb{R}^{I_1\times I_2\times ...\times I_N}$ is the number of linear components that could fit $\mathscr{X}$ exactly.
$
\mathscr{X}=\sum_{r=1}^R \textbf{a}_r^{(1)} \circ \textbf{a}_r^{(2)} \circ ... \circ \textbf{a}_r^{(N)} \;.
$

\subsection{Tensor Decomposition}
The CP decomposition, shown in figure \ref{fig::cptd}, has been introduced in \cite{harshman1970foundations,carroll1970analysis}. The tensor $\mathscr{X}\in\mathbb{R}^{I\times I\times K}$ is defined as a sum of rank-one tensors. The number of rank-one tensors is determined by the rank, denoted by $R$, of the tensor $\mathscr{X}$. The CP decomposition is expressed as 
$
  \mathscr{X} = \sum_{r=1}^{R} \textbf{a}_r^{(1)} \circ \textbf{a}_r^{(2)} \circ \textbf{a}_r^{(3)} \circ... \circ \textbf{a}_r^{(N)}
$
where $\textbf{a}_r^{(1)}, \textbf{a}_r^{(2)}, \textbf{a}_r^{(3)}, ..., \textbf{a}_r^{(N)}$ are factor vectors of size $\mathbb{R}^{I_1}, \mathbb{R}^{I_2}, \mathbb{R}^{I_3}, ..., \mathbb{R}^{I_N}$. Each factor vector $\textbf{a}_r^{(n)}$ with $n\in \left\lbrace 1, 2, ..., N \right\rbrace$ and $r \in \left\lbrace 1, ..., R \right\rbrace$ refers to one order and one rank of the tensor $\mathscr{X}$. \\

The DEDICOM decomposition \cite{harshman1978models}, illustrated in figure \ref{fig::dedicomtd}, describes the asymmetric relationships between $I$ objects of the tensor $\mathscr{X}\in \mathbb{R}^{I\times I \times K}$.  The decomposition groups the $I$ objects into $R$ latent components (or groups) and describes their pattern of interactions by computing $\textbf{A}\in\mathbb{R}^{I\times R}$, $\textbf{H}\in\mathbb{R}^{R\times R}$ and $\mathscr{D}\in\mathbb{R}^{R\times R \times K}$ such that
$
  \textbf{X}_k = \textbf{A}\textbf{D}_k \textbf{H} \textbf{D}_k \textbf{A}^T
  \quad , \quad \forall \: k = 1, ..., K .
$
The matrix $\textbf{A}$ indicates the participation of object $i = 1, ..., I$ in the group $r = 1, ..., R$, the matrix $\textbf{H}$ the interactions between the different components $r$ and the tensor $\mathscr{D}$ represents the participation of the $R$ latent component according to $K$.  \\

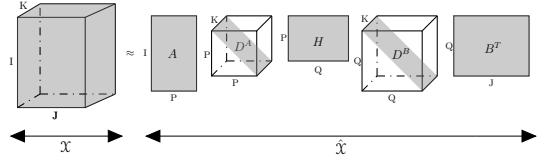
\begin{figure}[t]
  \centering
  \begin{tikzpicture}[scale=0.4, transform shape]

\draw (-7.25,-0.55) rectangle (-5,2.45);
\fill [gray, opacity=.4] (-7.25,-0.55) rectangle (-5,2.45);
\draw (-5,2.45) -- (-4,2.9);
\draw (-7.25,2.45) -- (-6.5,2.9);
\draw (-5,-0.55) -- (-4,-0.1);
\draw (-6.5,2.9) -- (-4,2.9);
\draw (-4,-0.1) -- (-4,2.9);
\draw [dashdotted] (-6.5,-0.1) -- (-4,-0.1);
\draw [dashdotted] (-6.5,-0.1) -- (-6.5,2.9);
\draw [dashdotted] (-7.25,-0.55) -- (-6.5,-0.1);

\fill [gray, opacity=.4] (-4,-0.1) -- (-4,2.9) -- (-5,2.45) -- (-5,-0.55)  -- cycle;
\fill [gray, opacity=.4] (-5,2.45) -- (-4,2.9) -- (-6.5,2.9) -- (-7.25,2.45)  -- cycle;

\node (none) at (-3.5,1.25) {$\approx$};

\draw  (-2.8,2.5) rectangle (-1.3,0);
\fill [gray, opacity=.4]  (-2.8,2.5) rectangle (-1.3,0);
\draw (-2.1,1.5) node[below]{\large{$A$}};

\draw (4.2,0) rectangle (6.2,2) node (v1) {};
\draw (6.2,2) -- (6.7,2.5);
\draw (4.2,2) -- (4.7,2.5);
\draw (6.2,0) -- (6.7,0.5);
\draw (4.7,2.5) -- (6.7,2.5);
\draw (6.7,0.5) -- (6.7,2.5);
\draw [dashdotted] (4.7,0.5) -- (6.7,0.5);
\draw [dashdotted] (4.7,0.5) -- (4.7,2.5);
\draw [dashdotted] (4.2,0) -- (4.7,0.5);
\fill [gray, opacity=.4] (6.2,0) -- (6.7,0.5) -- (4.7,2.5) -- (4.2,2)  -- cycle;
\draw (0.25,1.8) node[below]{\large{$D^A$}};

\draw (1.75,2.5) rectangle (3.75,1);
\fill [gray, opacity=.4] (1.75,2.5) rectangle (3.75,1);
\draw (2.75,1.9) node[below]{\large{$H$}};

\draw (-0.8,0.5) rectangle (0.7,2) node (v1) {};
\draw (0.7,2) -- (1.2,2.5);
\draw (-0.8,2) -- (-0.3,2.5);
\draw (0.7,0.5) -- (1.2,1);
\draw (-0.3,2.5) -- (1.2,2.5);
\draw (1.2,1) -- (1.2,2.5);
\draw [dashdotted] (-0.3,1) -- (1.2,1);
\draw [dashdotted] (-0.3,1) -- (-0.3,2.5);
\draw [dashdotted] (-0.8,0.5) -- (-0.3,1);
\fill [gray, opacity=.4] (0.7,0.5) -- (1.2,1) -- (-0.3,2.5) -- (-0.8,2)  -- cycle;
\draw (5.5,1.55) node[below]{\large{$D^B$}};

\draw  (7.25,2.5) rectangle (9.75,0.5);
\fill [gray, opacity=.4] (7.25,2.5) rectangle (9.75,0.5);
\draw (8.55,1.75) node[below]{\large{$B^T$}};

\draw [line width=0.25pt, -triangle 45] (-6.5,-1.5) -- (-7.5,-1.5);
\draw [line width=0.25pt, -triangle 45] (-6.5,-1.5) -- (-3.75,-1.5);

\draw [line width=0.25pt, -triangle 45] (8,-1.5) -- (-3,-1.5);
\draw [line width=0.25pt, -triangle 45] (8,-1.5) -- (10,-1.5);

\node (none) at (-5.65,-1.85) {\Large $\mathscr{X}$};
\node (none) at (3.5,-1.85) {\Large $\mathscr{\hat{X}}$};

\node (none) at (-7.45,1) {I};
\node (none) at (-6,-0.8) {J};
\node (none) at (-6,-0.8) {J};
\node (none) at (-7.05,2.85) {K};

\draw (-3,1.5) node[below]{\small{I}};
\draw (-2.05,0) node[below]{\small{P}};
\draw (0,0.5) node[below]{\small{P}};
\draw (-0.95,1.45) node[below]{\small{P}};
\draw (-0.7,2.6) node[below]{\small{K}};
\draw (4.3,2.65) node[below]{\small{K}};
\draw (1.6,2) node[below]{\small{P}};
\draw (2.75,0.95) node[below]{\small{Q}};
\draw (4.05,1.25) node[below]{\small{Q}};
\draw (5.2,0) node[below]{\small{Q}};
\draw (7.1,1.75) node[below]{\small{Q}};
\draw (8.5,0.5) node[below]{\small{J}};

\end{tikzpicture}
  \caption{Third order PARATUCK2 TD}
  \label{fig::paratuck2td}
\end{figure}

The PARATUCK2 decomposition \cite{harshman1996uniqueness}, presented in figure \ref{fig::paratuck2td}, expresses the original tensor $\mathscr{X} \in \mathbb{R}^{I\times J\times K}$ as a product of matrices and tensors
$
\textbf{X}_k = \textbf{A}\textbf{D}^A_k\textbf{H}\textbf{D}^B_k\textbf{B}^T
$ 
with 
$
k=\left\lbrace 1, ..., K \right\rbrace
$
where $\textbf{A}$, $\textbf{H}$ and $\textbf{B}$ are matrices of size $\mathbb{R}^{I\times P}$, $\mathbb{R}^{P \times Q}$, and $\mathbb{R}^{J\times Q}$. The matrices $\textbf{D}^A_k\in \mathbb{R}^{P\times P}$ and $\textbf{D}^B_k\in \mathbb{R}^{Q\times Q}\:\forall k\in\left\lbrace 1, ...,K \right\rbrace$ are the slices of the tensors $\mathscr{D}^A\in \mathbb{R}^{P\times P \times K}$ and $\mathscr{D}^B\in \mathbb{R}^{Q\times Q \times K}$. The columns of the matrices $\textbf{A}$ and $\textbf{B}$ represent the latent factors $P$ and $Q$, and therefore the rank of each object set. The matrix $\textbf{H}$ underlines the asymmetry between the $P$ latent components and the $Q$ latent components. The tensors $\mathscr{D}^A$ and $\mathscr{D}^B$ measure the evolution of the latent components according to $K$.

\begin{figure}[t]
\begin{minipage}{0.4\linewidth} 
  \centering
  \begin{tikzpicture}[scale=0.4, transform shape]
\draw (-7.25,-0.55) rectangle (-5,2.45);
\fill [gray, opacity=.4] (-7.25,-0.55) rectangle (-5,2.45);
\draw (-5,2.45) -- (-4,2.9);
\draw (-7.25,2.45) -- (-6.5,2.9);
\draw (-5,-0.55) -- (-4,-0.1);
\draw (-6.5,2.9) -- (-4,2.9);
\draw (-4,-0.1) -- (-4,2.9);
\draw [dashdotted] (-6.5,-0.1) -- (-4,-0.1);
\draw [dashdotted] (-6.5,-0.1) -- (-6.5,2.9);
\draw [dashdotted] (-7.25,-0.55) -- (-6.5,-0.1);

\fill [gray, opacity=.4] (-4,-0.1) -- (-4,2.9) -- (-5,2.45) -- (-5,-0.55)  -- cycle;
\fill [gray, opacity=.4] (-5,2.45) -- (-4,2.9) -- (-6.5,2.9) -- (-7.25,2.45)  -- cycle;

\node (none) at (-3.25,1.75) {$\approx$};

\coordinate (A) at (-2.65,-0.2) {} {} {} {} {} {};
\coordinate (B) at (-2.3,1.8) {} {} {} {} {} {};
\coordinate (C) at (-1.95,1.65) {} {} {} {} {} {};
\coordinate (D) at (-0.45,2) {} {} {} {} {} {};
\coordinate (e1) at (-2.4,2.2) {} {} {} {} {} {};
\coordinate (e2) at (-1.95,2.2) {} {} {} {} {} {};
\coordinate (e3) at (-0.95,2.7) {} {} {} {} {} {};
\coordinate (e4) at (-1.4,2.7) {} {} {} {} {} {};
\draw (A) rectangle (B);
\fill [gray, opacity=.4] (A) rectangle (B);
\draw (C) rectangle (D);
\fill [gray, opacity=.4] (C) rectangle (D);
\draw (e1) -- (e2) -- (e3) -- (e4) -- cycle;
\fill [gray, opacity=.4] (e1) -- (e2) -- (e3) -- (e4) -- cycle;

\node (none) at (0.05,1.85) {$+$};
\node (none) at (0.55,1.85) {$...$};
\node (none) at (1.05,1.85) {$+$};

\coordinate (A) at (1.6,-0.2) {} {} {} {} {} {} {} {};
\coordinate (B) at (1.95,1.8) {} {} {} {} {} {} {} {};
\coordinate (C) at (2.3,1.65) {} {} {} {} {} {} {} {};
\coordinate (D) at (3.8,2) {} {} {} {} {} {} {} {};
\coordinate (e1) at (1.85,2.2) {} {} {} {} {} {} {} {};
\coordinate (e2) at (2.3,2.2) {} {} {} {} {} {} {} {};
\coordinate (e3) at (3.3,2.7) {} {} {} {} {} {} {} {};
\coordinate (e4) at (2.85,2.7) {} {} {} {} {} {} {} {};
\draw (A) rectangle (B);
\fill [gray, opacity=.4] (A) rectangle (B);
\draw (C) rectangle (D);
\fill [gray, opacity=.4] (C) rectangle (D);
\draw (e1) -- (e2) -- (e3) -- (e4) -- cycle;
\fill [gray, opacity=.4] (e1) -- (e2) -- (e3) -- (e4) -- cycle;

\draw [line width=0.25pt, -triangle 45] (-6,-1.5) -- (-3.75,-1.5);
\draw [line width=0.25pt, -triangle 45] (-6,-1.5) -- (-7.5,-1.5);

\draw [line width=0.25pt, -triangle 45] (3,-1.5) -- (-3,-1.5);
\draw [line width=0.25pt, -triangle 45] (3,-1.5) -- (4.0,-1.5);

\node (none) at (-5.5,-1.85) {\Large $\mathscr{X}$};
\node (none) at (0.5,-1.85) {\Large $\mathscr{\hat{X}}$};
\node (none) at (-7.45,1) {\small{I}};
\node (none) at (-6,-0.8) {\small{J}};
\node (none) at (-7.05,2.85) {\small{K}};
\node (none) at (-2.45,-0.6) {$a_1^{(1)}$};
\node (none) at (-1.15,1.3) {$a_1^{(2)}$};
\node (none) at (-2.1,2.75) {$a_1^{(3)}$};
\node (none) at (1.8,-0.6) {$a_R^{(1)}$};
\node (none) at (3.05,1.3) {$a_R^{(2)}$};
\node (none) at (2.2,2.75) {$a_R^{(3)}$};

\end{tikzpicture}
  \caption{Third order CP TD}
  \label{fig::cptd}
\end{minipage}
\hfill
\begin{minipage}{0.55\linewidth} 
  \centering
  \begin{tikzpicture}[scale=0.375, transform shape]

\draw (-7.25,-0.55) rectangle (-5,2.45);
\fill [gray, opacity=.4] (-7.25,-0.55) rectangle (-5,2.45);
\draw (-5,2.45) -- (-4,2.9);
\draw (-7.25,2.45) -- (-6.5,2.9);
\draw (-5,-0.55) -- (-4,-0.1);
\draw (-6.5,2.9) -- (-4,2.9);
\draw (-4,-0.1) -- (-4,2.9);
\draw [dashdotted] (-6.5,-0.1) -- (-4,-0.1);
\draw [dashdotted] (-6.5,-0.1) -- (-6.5,2.9);
\draw [dashdotted] (-7.25,-0.55) -- (-6.5,-0.1);

\fill [gray, opacity=.4] (-4,-0.1) -- (-4,2.9) -- (-5,2.45) -- (-5,-0.55)  -- cycle;
\fill [gray, opacity=.4] (-5,2.45) -- (-4,2.9) -- (-6.5,2.9) -- (-7.25,2.45)  -- cycle;

\node (none) at (-3.5,1.25) {$\approx$};

\draw  (-2.8,2.5) rectangle (-1.3,0);
\fill [gray, opacity=.4]  (-2.8,2.5) rectangle (-1.3,0);
\draw (-2.1,1.5) node[below]{\large{$A$}};

\draw (4.25,0.5) rectangle (5.75,2) node (v1) {};
\draw (5.75,2) -- (6.25,2.5);
\draw (4.25,2) -- (4.75,2.5);
\draw (5.75,0.5) -- (6.25,1);
\draw (4.75,2.5) -- (6.25,2.5);
\draw (6.25,1) -- (6.25,2.5);
\draw [dashdotted] (4.75,1) -- (6.25,1);
\draw [dashdotted] (4.75,1) -- (4.75,2.5);
\draw [dashdotted] (4.25,0.5) -- (4.75,1);
\fill [gray, opacity=.4] (5.75,0.5) -- (6.25,1) -- (4.75,2.5) -- (4.25,2)  -- cycle;
\draw (0.2,1.75) node[below]{\large{$D$}};

\draw (1.75,2.5) rectangle (3.75,1);
\fill [gray, opacity=.4] (1.75,2.5) rectangle (3.75,1);
\draw (2.75,1.9) node[below]{\large{$H$}};

\draw (-0.8,0.5) rectangle (0.7,2) node (v1) {};
\draw (0.7,2) -- (1.2,2.5);
\draw (-0.8,2) -- (-0.3,2.5);
\draw (0.7,0.5) -- (1.2,1);
\draw (-0.3,2.5) -- (1.2,2.5);
\draw (1.2,1) -- (1.2,2.5);
\draw [dashdotted] (-0.3,1) -- (1.2,1);
\draw [dashdotted] (-0.3,1) -- (-0.3,2.5);
\draw [dashdotted] (-0.8,0.5) -- (-0.3,1);
\fill [gray, opacity=.4] (0.7,0.5) -- (1.2,1) -- (-0.3,2.5) -- (-0.8,2)  -- cycle;
\draw (5.2,1.7) node[below]{\large{$D$}};

\draw  (6.7,2.5) rectangle (9.2,1);
\fill [gray, opacity=.4] (6.7,2.5) rectangle (9.2,1);
\draw (7.95,2) node[below]{\large{$A^T$}};

\draw [line width=0.25pt, -triangle 45] (-6.5,-1.5) -- (-7.5,-1.5);
\draw [line width=0.25pt, -triangle 45] (-6.5,-1.5) -- (-3.75,-1.5);

\draw [line width=0.25pt, -triangle 45] (8,-1.5) -- (-3,-1.5);
\draw [line width=0.25pt, -triangle 45] (8,-1.5) -- (9.25,-1.5);

\node (none) at (-5.65,-1.85) {\Large $\mathscr{X}$};
\node (none) at (3,-1.85) {\Large $\mathscr{\hat{X}}$};

\node (none) at (-7.45,1) {\small{I}};
\node (none) at (-6,-0.8) {\small{I}};
\node (none) at (-7.05,2.85) {\small{K}};

\draw (-3,1.5) node[below]{\small{I}};
\draw (-2.05,0) node[below]{\small{P}};
\draw (0,0.5) node[below]{\small{P}};
\draw (-0.95,1.55) node[below]{\small{P}};
\draw (-0.7,2.6) node[below]{\small{K}};
\draw (4.35,2.65) node[below]{\small{K}};
\draw (1.6,1.95) node[below]{\small{P}};
\draw (2.75,0.95) node[below]{\small{P}};
\draw (4.1,1.5) node[below]{\small{P}};
\draw (5,0.5) node[below]{\small{P}};
\draw (6.55,1.9) node[below]{\small{P}};
\draw (7.95,1) node[below]{\small{I}};

\end{tikzpicture}
  \caption{Third order DEDICOM TD}
  \label{fig::dedicomtd}
\end{minipage}
\end{figure}

\section{VecHGrad for Tensor Decomposition} \label{sec::propmethod}
In this section, we first introduce VecHGrad for the first order tensors, also commonly called vectors. We then present the core of our contribution, VecHGrad for N-order tensor decomposition. 

\subsection{Introduction to VecHGrad for Vectors}
Under Newton's method, the current iterate $\Tilde{\textbf{x}}^t \in \mathcal{C}$ is used to generate the next iterate $\Tilde{\textbf{x}}^{t+1}$ by performing a constrained minimization of the second order Taylor expansion such that:

\begin{equation}
\setlength{\abovedisplayskip}{0pt}
\begin{split}
  \Tilde{\textbf{x}}^{t+1} = \arg \underset{\textbf{x}\in \mathcal{C}}{\min} 
  \{ &
  \dfrac{1}{2} \left \langle \textbf{x}-\Tilde{\textbf{x}}^t, \nabla^2 f(\Tilde{\textbf{x}}^t)(\textbf{x}-\Tilde{\textbf{x}}^t)  \right \rangle 
  + \left \langle \nabla f(\Tilde{\textbf{x}}^t), \textbf{x}-\Tilde{\textbf{x}}^t \right \rangle
  \}  
\end{split}  \; .
\end{equation}

We recall that $\nabla f$ and $\nabla^2 f$ denote the gradient and the Hessian matrix, respectively, of the objective function $f : \mathbb{R}^d \rightarrow \mathbb{R}$. 

\begin{minipage}{0.4\linewidth} 
\begin{equation}
\begin{split}
  \nabla f = \;& \underset{\textbf{x}\in\mathcal{C}}{grad} f \\
  \nabla f = \;& [\dfrac{\partial f}{\partial x_1}, \dfrac{\partial f}{\partial x_2}, ..., 
  \dfrac{\partial f}{\partial x_d}]
\end{split}
\end{equation}
\end{minipage}
\hspace{.5cm}
\begin{minipage}{0.55\linewidth} 
\begin{equation} \label{eq::Hessian_matrix}
\begin{split}
\nabla^2 f = \;& \textbf{Hes} \\
\nabla^2 f = \;& 
\begin{pmatrix}
\frac{\partial^2 f}{\partial x_1^2} & 
\frac{\partial^2 f}{\partial x_1 \partial x_2} & 
\cdots & 
\frac{\partial^2 f}{\partial x_1 \partial x_d} \\
\vdots & \vdots & \ddots & \vdots \\[.25em]
\frac{\partial^2 f}{\partial x_d \partial x_1} &
\frac{\partial^2 f}{\partial x_d \partial x_2} & 
\cdots & 
\frac{\partial^2 f}{\partial x_d^2}
\end{pmatrix}
\end{split}
\end{equation}
\end{minipage}



\vspace{.2cm}
When $\mathcal{C}\in\mathbb{R}^d$, which is the unconstrained form, the new iterate $\Tilde{\textbf{x}}^{t+1}$ is generated such that:
\begin{equation}
\setlength{\abovedisplayskip}{0pt}
  \Tilde{\textbf{x}}^{t+1} = \Tilde{\textbf{x}}^t - [\nabla^2f(\Tilde{\textbf{x}}^t)]^{-1} \nabla f(\Tilde{\textbf{x}}^t) \; .
\end{equation}

We use the strong Wolfe's line search which allows Newton's method to be globally convergent. The line search is defined by the following three inequalities:

\begin{equation}
\setlength{\abovedisplayskip}{0pt}
\begin{split}
 \text{i) }& f(\Tilde{\textbf{x}}^t+\alpha^t \textbf{p}^t) \leq f(\Tilde{\textbf{x}}^t) + c_1 \alpha^t (\textbf{p}^t)^T \nabla f (\Tilde{\textbf{x}}^t) \; , \\ 
 \text{ii) }& -(\textbf{p}^t)^T \nabla f(\Tilde{\textbf{x}}^t + \alpha^t \textbf{p}^t) \leq -c_2 (\textbf{p}^t)^T \nabla f(\Tilde{\textbf{x}}^t) \; , \\ 
 \text{iii) }& \mid (\textbf{p}^t)^T \nabla f(\Tilde{\textbf{x}}^t + \alpha^t \textbf{p}^t) \mid \leq c_2 \mid (\textbf{p}^t)^T \nabla f(\Tilde{\textbf{x}}^t) \mid \; ,
\end{split}
\end{equation}

where $0\leq c_1 \leq c_2 \leq 1$, $\alpha^t > 0$ is the step length and $\textbf{p}^t = -[\nabla^2 f(\Tilde{\textbf{x}}^t)]^{-1}\nabla f(\Tilde{\textbf{x}}^t)$. Therefore, the iterate $\Tilde{\textbf{x}}^{t+1}$ becomes the following:

\begin{equation}
\setlength{\abovedisplayskip}{0pt}
\begin{dcases}
   \Tilde{\textbf{x}}^{t+1} = & \Tilde{\textbf{x}}^t - \alpha^t [\nabla^2f(\Tilde{\textbf{x}}^t)]^{-1} \nabla f(\Tilde{\textbf{x}}^t) \; ,\\
   \Tilde{\textbf{x}}^{t+1} = & \Tilde{\textbf{x}}^t + \alpha^t \textbf{p}^t \; .
\end{dcases}
\end{equation}

Computing the inverse of the exact Hessian matrix, $[\nabla^2f(\Tilde{\textbf{x}}^t)]^{-1}$, may be difficult. The inverse is therefore computed with a Conjugate Gradient (CG) loop. It has two main advantages: the calculations are considerably less expensive \cite{wright1999numerical} 
and the Hessian can be expressed by a diagonal approximation. The convergence of the CG loop is defined when a maximum number of iterations is reached or when the residual $\textbf{r}^t = \nabla^2 f(\Tilde{\textbf{x}}^t)\textbf{p}^t + \nabla f(\Tilde{\textbf{x}}^t)$ satisfies $\parallel \textbf{r}^t \parallel \leq \sigma \parallel \nabla f(\Tilde{\textbf{x}}^t) \parallel$ with $\sigma \in \mathbb{R}^+$. In the CG loop, the exact Hessian matrix is approximated by a diagonal approximation. The Hessian matrix is multiplied by a descent direction vector resulting in a vector 
which satisfies the requirement of the main optimization loop. Therefore, only the results of the Hessian vector product are needed. The equation is expressed below: 

\begin{equation} \label{eq::hesvectdot1}
\nabla^2 f(\Tilde{\textbf{x}}^t) \: \textbf{p}^t = \dfrac{\nabla f(\Tilde{\textbf{x}}^t+\eta  \: \textbf{p}^t)-\nabla f(\Tilde{\textbf{x}}^t)}{\eta} \; .
\end{equation}

The term $\eta$ is the perturbation and the term $\textbf{p}^t$ is the descent direction vector, set to the negative of the gradient at initialization. Consequently, the extensive computation of the inverse of the exact full Hessian matrix is bypassed using only gradient diagonal approximation. 





  
  
  


\subsection{VecHGrad for Accurate Resolution of Tensor Decomposition}


The loss function, or the objective function, is denoted by $f$. It is equal to: 

\begin{equation} \label{eq::errorminimization}
  f(\Tilde{\textbf{x}}) = \min_{\mathscr{\hat{X}}} ||\mathscr{X}-\mathscr{\hat{X}}|| .
\end{equation}
The tensor $\mathscr{X}$ is the original tensor and the tensor $\mathscr{\hat{X}}$ is the approximated tensor built from the decomposition. If we consider, for instance, that the CP TD applied on a third order tensor, the tensor $\mathscr{\hat{X}}$ is the tensor built with the factor vectors $\textbf{a}_r^{(1)}, \textbf{a}_r^{(2)}, \textbf{a}_r^{(3)}$ for $r=1,...,R$ initially randomized such as:

\begin{equation}
  \mathscr{\hat{X}} = \sum_{r=1}^{R} \textbf{a}_r^{(1)} \circ \textbf{a}_r^{(2)} \circ \textbf{a}_r^{(3)} .
\end{equation}

The vector $\Tilde{\textbf{x}}$ is a flattened vector containing all the entries of the decomposed tensor $\mathscr{\hat{X}}$. If we consider a third order tensor $\mathscr{\hat{X}}$ of rank $R$ factorized with the CP TD, we obtain the following vector $\Tilde{\textbf{x}}\in\mathbb{R}^{d=R(I + J + K)}$ such that:

\begin{equation} \label{eq::vectorize_x}
\begin{split}
  \Tilde{\textbf{x}} = \: \text{vec}(\mathscr{\hat{X}}) 
  =
  [
    & \textbf{a}^{(1)}_{1}, \; \textbf{a}^{(1)}_{2}, ..., \textbf{a}^{(R)}_{I} , 
    \textbf{a}^{(2)}_{1}, \; \textbf{a}^{(2)}_{2}, ..., \textbf{a}^{(R)}_{J} , 
    \textbf{a}^{(3)}_{1}, \; \textbf{a}^{(3)}_{2}, ..., \textbf{a}^{(R)}_{K}
  ]^T
\end{split} \;.
\end{equation}

Since the objective is to propose a universal approach for any TDs, we rely on finite difference method to compute the gradient of the loss function of any TDs. Thus, the method can be transposed to any decomposition just by changing the decomposition equation. The approximate gradient is based on a fourth order formula (\ref{eq::fourth_order_diff}) to ensure reliable approximation \cite{schittkowski2001nlpqlp}:

\begin{equation} \label{eq::fourth_order_diff}
\begin{split}
\dfrac{\partial}{\partial x_i} f(\Tilde{\textbf{x}}^t) 
\approx &
\dfrac{1}{4!\eta}  \big( 
2 [f(\Tilde{\textbf{x}}^t-2\eta \textbf{e}_i) - f(\Tilde{\textbf{x}}^t + 2 \eta \textbf{e}_i)] \\
& \quad + 16 [f(\Tilde{\textbf{x}}^t + \eta \textbf{e}_i) - f(\Tilde{\textbf{x}}^t - \eta \textbf{e}_i)]
\big)
\end{split} \; .
\end{equation}

In Equation (\ref{eq::fourth_order_diff}), the index $i$ is the index of the variables for which the derivative is to be evaluated. The variable $\textbf{e}_i$ is the $i$th unit vector. The term $\eta$, the perturbation, is set to a small value to achieve the convergence of the 
process.  \\

The Hessian diagonal approximation is evaluated as described in the previous section, using Equation (\ref{eq::hesvectdot1}). Our approach is therefore free of the extensive computation of the inverse of the exact Hessian matrix.
%
%
We finally reached the core  objective of describing VecHGrad for tensors, summarized in Algorithm \ref{algo:vechgrad}.






\section{Experiments} \label{sec::exp}
We investigate the convergence behavior of VecHGrad and compare it to other popular optimizers inherited from both the tensor and machine learning communities. We compare VecHGrad with ten different algorithms applied to the three main TDs, CP, DEDICOM, and PARATUCK2, with increasing linear algebra complexity: 
\begin{itemize}
\item ALS, Alternating Least Squares \cite{bro1998multi,charlier2018non};
\item SGD, Stochastic Gradient Descent \cite{wright1999numerical};
\item NAG, Nesterov Accelerated Gradient \cite{nesterov2007gradient};
\item Adam \cite{kingma2014adam};
\item RMSProp \cite{hinton2012rmsprop};
\item SAGA \cite{defazio2014saga};
\item AdaGrad \cite{duchi2011adaptive};
\item CP-OPT and the Non-linear Conjugate Gradient (NCG) \cite{acar2011scalable,acar2011all};
\item L-BFGS \cite{liu1989limited}. 
\end{itemize}

\SetAlFnt{\scriptsize}
\SetAlCapFnt{\scriptsize}
\SetAlCapNameFnt{\scriptsize}

\begin{algorithm}[t]
\setstretch{1.25}
\DontPrintSemicolon

Select tensor decomposition equation:
    $g: \left\{\begin{matrix*}[l]
  \mathbb{R}^d \rightarrow  \mathbb{R}^{I_1\times I_2\times ...\times I_n}  \\ 
  \Tilde{\textbf{x}}^t \mapsto \mathscr{\Tilde{X}} 
  \end{matrix*}\right.$ \\
Select randomly $\Tilde{\textbf{x}}^0 \:=\: \text{vec}(\mathscr{\hat{X}})$  \\
\Repeat{$t = \text{maxiter} \:$ 
  \textsc{or} $\: f(\Tilde{\textbf{x}}^t)\leq \epsilon_1\:$ 
  \textsc{or} $\: \parallel \nabla f(\Tilde{\textbf{x}}^t) \parallel \leq \epsilon_2$}{
  Select loss function: 
  $f: \left\{\begin{matrix*}[l]
  \mathbb{R}^d \rightarrow \mathbb{R} \\ 
  \Tilde{\textbf{x}}^t \mapsto \parallel \mathscr{X} - g(\Tilde{\textbf{x}}^t) \parallel
  \end{matrix*}\right. $\\
  
  Compute gradient $\nabla f(\Tilde{\textbf{x}}^t) \in \mathbb{R}^d$

  Fix $\textbf{p}_0^t = - \nabla f(\Tilde{\textbf{x}}^t)$
  
  \Repeat{$k = \textit{cg}_{\text{maxiter}} \:$ \textsc{or} $\: \parallel \textbf{r}_k \parallel \leq \sigma \parallel \nabla f(\Tilde{\textbf{x}})^t) \parallel$}{
  Update $\textbf{p}_k^t$ with CG loop:
      $\textbf{r}_k = \nabla^2 f(\Tilde{\textbf{x}}^t)\textbf{p}_k^t + \nabla f(\Tilde{\textbf{x}}^t)$ \\
  }
  
  $\alpha^t \leftarrow $ Wolfe's line search
  
  Update parameters:
      $\Tilde{\textbf{x}}^{t+1} = \Tilde{\textbf{x}}^t + \alpha^t \textbf{p}^t_{\text{opt}}$  \\
  
}

\caption{VecHGrad, tensor case \label{algo:vechgrad}}
\end{algorithm}

\textbf{Data Availability and Code Availability}
We highlight the performance of VecHGrad using 5 popular data sets CIFAR10, CIFAR100, MNIST, LFW and COCO, all available online. 
Each data set has different intrinsic characteristics such as size or sparsity. A quick overview of their characteristics is presented in Table \ref{tab::dataset}. We chose to use different data sets as the performance of different optimizers might vary depending on the data. The overall conclusion for the experiments is therefore independent of one particular data set. 
The implementation and the data for the experiments are available on GitHub\footnote{\label{note1}The code is available at https://github.com/dagrate/vechgrad.}.  \\

\begin{table}[t]
  \caption{Description of the data sets used (K: thousands).}
  \label{tab::dataset}
  \centering
  \scalebox{0.65}{
    \setlength{\tabcolsep}{10pt}
    \begin{tabular}{cccc}
	\toprule
    Data Set & Labels & Size & Batch Size\\
    \midrule
    CIFAR-10 & image $\times$ pixels $\times$ pixels & 50K $\times$ 32 $\times$ 32 & 64 \\
    CIFAR-100 & image $\times$ pixels $\times$ pixels & 50K $\times$ 32 $\times$ 32 & 64 \\
    MNIST & image $\times$ pixels $\times$ pixels & 60K $\times$ 28 $\times$ 28 & 64 \\
    COCO & image $\times$ pixels $\times$ pixels & 123K $\times$ 64 $\times$ 64 & 32 \\
    LFW & image $\times$ pixels $\times$ pixels & 13K $\times$ 64 $\times$ 64 & 32 \\
    \bottomrule
  \end{tabular}
  }
\end{table}

\textbf{Experimental Setup}
In our experiments, we use the standard parameters for the popular ML and DL gradient optimization methods. We use $\eta = 10^{-4}$ for SGD, $\gamma = 0.9$ and $\eta = 10^{-4}$ for NAG, $\beta_1 = 0.9, \beta_2 = 0.999, \epsilon = 10^{-8}$ and $\eta = 0.001$ for Adam, $\gamma = 0.9, \eta = 0.001$ and $\epsilon = 10^{-8}$ for RMSProp, $\eta = 10^{-4}$ for SAGA, $\eta = 0.01$ and $\epsilon = 10^{-8}$ for AdaGrad. We use the Hestenes-Stiefel update for the NCG resolution. Furthermore, the convergence criteria is reached when $f^{i+1} - f^i \leq 0.001$ or when a maximum number of iterations is reached. We use 100,000 iterations for gradient-free methods, 10,000 iterations for gradient methods and 1,000 iterations for Hessian-based methods. Additionally, we fixed the number of iterations to 20 for VecHGrad's inner CG loop, used to determine the descent direction. We invite the reader to see our code available on GitHub\textsuperscript{\ref{note1}} for further insights about the parameters used in our study. The simulations were conducted on a server with 50 Intel Xeon E5-4650 CPU cores and 50GB of RAM. All the resolution schemes have been implemented in Julia and are compatible with the ArrayFire GPU accelerator library.  \\

\begin{figure}[t]
  \centering
  \includegraphics[scale=.2]{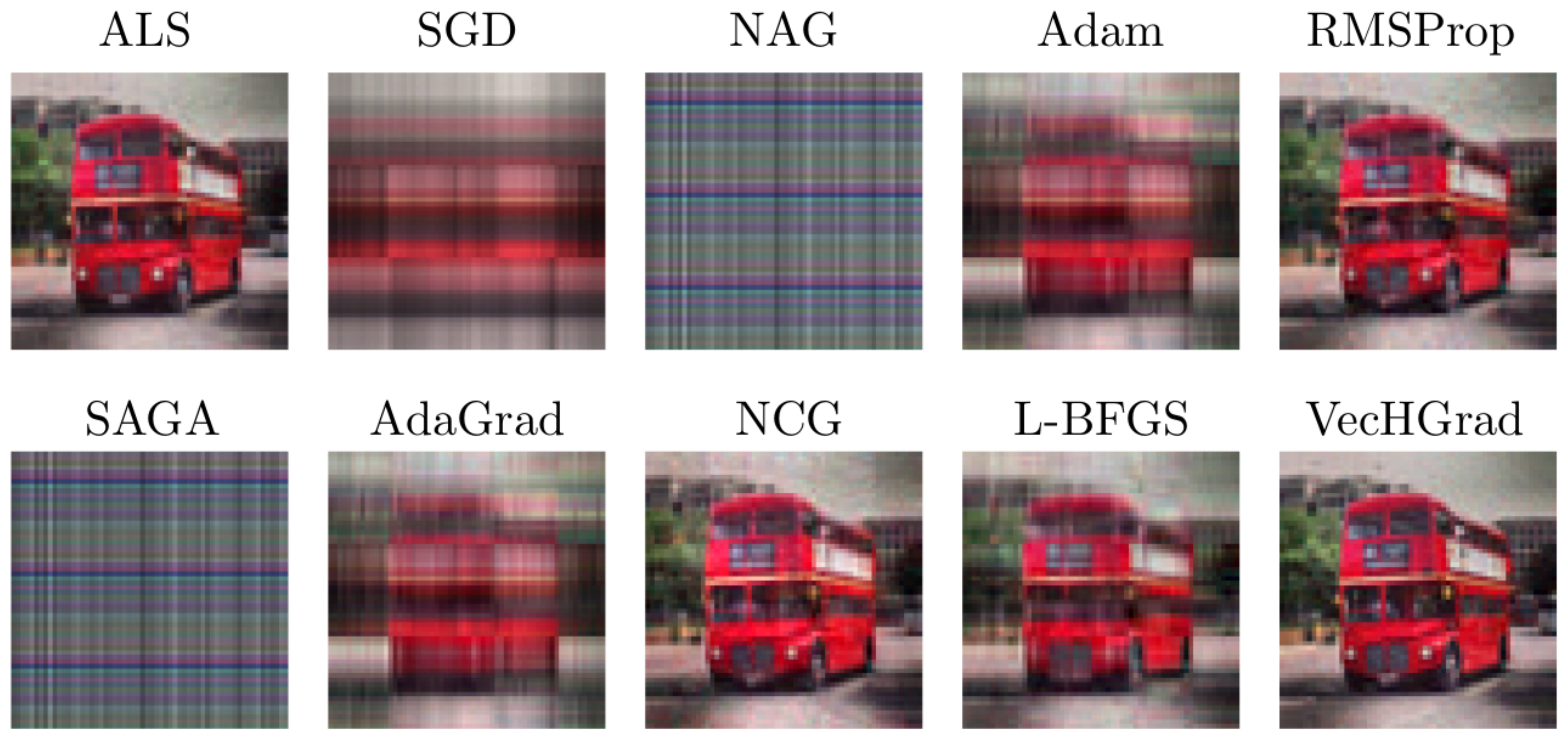}
  \caption{Visual simulation of the accuracy at convergence of the different optimizers for the PARATUCK2 decomposition. The accuracy is highlighted by how blurry the image is (error at convergence ALS: 1018, SGD: 4082, NAG: 6469, Adam: 2996, RMSProp: 1184, SAGA: 6378, AdaGrad: 2961, NCG: 1569, L-BFGS: 1771, VecHGrad: 599). The popular gradient optimizers AdaGrad, NAG and SAGA failed to converge to a solution close to the original image, contrarily to VecHGrad or RMSProp.} 
  \label{fig::visualplot}
\end{figure}

\textbf{Results and Discussion}
%
First, we performed an experiment to identify visually the strengths of each of the optimization algorithms. The figure \ref{fig::visualplot} depicts the resulting error of the loss function for each of the methods at convergence for PARATUCK2 TD. We fixed latent components for which the differences between the optimizers are easily noticeable. 
The error of the loss function 
depends on how blurry the picture is. The less the image is blurry, the lower the loss function error at convergence. Some of the numerical methods, ALS, RMSProp and VecHGrad, offered the best observable image quality at convergence, given our choice of parameters. Other popular schemes, including NAG and SAGA, however failed to converge to a solution resulting in a noisy image, far from being close to the original one.  \\

In a second experiment, we compared in Table \ref{tab::accuracyparatuck2} and Table \ref{tab::timeparatuck2} the loss function errors and the calculation times of the optimizers for the five ML data sets described in Table \ref{tab::dataset} for the three TDs aforementioned. Both the loss function errors and the calculation times were computed based on the mean of the loss function errors and the mean of the calculation times over all batches. The numerical schemes of the NAG, SAGA and AdaGrad algorithms failed to minimize the error of the loss function accurately. The ALS scheme furthermore offers a good compromise between the resulting errors and the required calculation times, explaining its major success across TD applications. The gradient descent optimizers, Adam and RMSProp, and the Hessian based optimizers, VechGrad and L-BFGS, were capable to minimize the loss function the most accurately. The NCG method achieved satisfying errors for the CP and the DEDICOM decomposition but its performance decreases significantly when trying to solve the PARATUCK2 decomposition. Surprisingly, the calculation times of the Adam and RMSProp gradient descents were greater than the calculation times of VecHGrad. VecHGrad was capable to outperform the gradient descent schemes for both accuracy and speed thanks to the use of the strong Wolfe's line search and the vector Hessian approximation, inherited from gradient information. This result is reported in Table \ref{tab::conv-rate} presenting the experimental mean convergence rate, defined such that
$ q \approx 
 \log \left | \frac{\textbf{f}^{t+1}-\textbf{f}^{t}}{\textbf{f}^{t}-\textbf{f}^{t-1}} \right |
 \begin{bmatrix}
   \log \left | \frac{\textbf{f}^{t}-\textbf{f}^{t-1}}{\textbf{f}^{t-1}-\textbf{f}^{t-2}} \right |
 \end{bmatrix}^{-1}$ 
for PARATUCK2 TD. The latter underlines the best differences between the optimizers. The highest convergence rate was obtained by VecHGrad, followed by L-BFGS and NCG. Similar values were obtained for the other decomposition. Thus, we can conclude that VecHGrad is capable to solve accurately and efficiently TDs outperforming popular machine learning gradient descent algorithms.  \\

\begin{table}[p]
\centering
\caption{Mean of the loss function errors at convergence over all batches. The lower, the better. 
The strong Wolfe's line search is crucial for the VecHGrad's performance.}
\label{tab::accuracyparatuck2}
\scalebox{0.6}{
\setlength{\tabcolsep}{6.25pt}
\begin{tabular}{ccccccccccc}
\toprule
  Decomposition & Optimizer & CIFAR-10 & CIFAR-100 & MNIST & COCO & LFW \\
\midrule
  CP & ALS & 318.667 & 428.402 & 897.766 & 485.138 & 4792.605 \\
  CP & SGD & 2112.904 & 2825.710 & 2995.528 & 3407.415 & 7599.458 \\
  CP & NAG & 4338.492 & 5511.272 & 4916.003 & 8187.315 & 18316.589 \\
  CP & Adam & 1578.225 & 2451.217 & 1631.367 & 2223.211 & 6644.167 \\
  CP & RMSProp & 127.961 & 128.137 & 200.002 & 86.792 & 4205.520 \\
  CP & SAGA & 4332.879 & 5501.528 & 4342.708 & 6327.580 & 13242.181 \\
  CP & AdaGrad & 3142.583 & 4072.551 & 2944.768 & 4921.861 & 10652.488 \\
  CP & NCG & 41.990 & 37.086 & 23.320 & 76.478 & 4130.942 \\
  CP & L-BFGS & 195.298 & 525.279 & 184.906 & 596.160 & 4893.815 \\
  CP & VecHGrad & \textbf{$\leq$ 1.000} & \textbf{$\leq$ 1.000} & \textbf{$\leq$ 1.000} & \textbf{$\leq$ 1.000} & \textbf{$\leq$ 1.000} \\[.15cm]

DEDICOM & ALS & 1350.991 & 	1763.718 & 	1830.830 & 	1894.742 & 	3193.685  \\
DEDICOM & SGD & 435.780 & 	456.051 & 	567.503 & 	406.760 & 	511.093   \\
DEDICOM & NAG & 4349.151 & 	5722.073 & 	4415.687 & 	6325.638 & 	9860.454  \\
DEDICOM & Adam & 579.723 & 	673.316 & 	575.341 & 	743.977 & 	541.515   \\
DEDICOM & RMSProp & 63.795 & 	236.974 & 	96.240 & 	177.419 & 	33.224    \\
DEDICOM & SAGA & 4285.512 & 	5577.981 & 	4214.771 & 	5797.562 & 	8128.724  \\ 
DEDICOM & AdaGrad & 1962.966 & 	2544.436 & 	1452.278 & 	2851.649 & 	3033.965  \\ 
DEDICOM & NCG & 550.554 & 	321.332 & 	171.181 & 	583.430 & 	711.549   \\
DEDICOM & L-BFGS & 423.802 & 	561.689 & 	339.284 & 	435.188 & 	511.620   \\
DEDICOM & VecHGrad & \textbf{$\leq$ 1.000} & \textbf{$\leq$ 1.000} & \textbf{$\leq$ 1.000} & \textbf{$\leq$ 1.000} & \textbf{$\leq$ 1.000} \\[.15cm]

PARATUCK2 & ALS & 408.724 & 	480.312 & 	1028.250 & 	714.623 & 	658.284   \\ 
PARATUCK2 & SGD & 639.556 & 	631.870 & 	1306.869 & 	648.962 & 	495.188   \\
PARATUCK2 & NAG & 4699.058 & 	6046.024 & 	5168.824 & 	8205.223 & 	14546.438 \\ 
PARATUCK2 & Adam & 512.725 & 	680.653 & 	591.156 & 	594.687 & 	615.731   \\ 
PARATUCK2 & RMSProp & 133.416 & 	145.766 & 	164.709 & 	134.047 & 	174.769   \\ 
PARATUCK2 & SAGA & 4665.435 & 	5923.178 & 	4934.328 & 	6350.172 & 	8847.886  \\ 
PARATUCK2 & AdaGrad & 1775.433 & 	2310.402 & 	1715.316 & 	2752.348 & 	2986.919  \\ 
PARATUCK2 & NCG & 772.634 & 	1013.032 & 	270.288 & 	335.532 & 	15181.961 \\ 
PARATUCK2 & L-BFGS & 409.666 & 	522.158 & 	464.259 & 	467.139 & 	416.761   \\ 
PARATUCK2 & VecHGrad & \textbf{$\leq$ 1.000} & \textbf{$\leq$ 1.000} & \textbf{$\leq$ 1.000} & \textbf{$\leq$ 1.000} & \textbf{$\leq$ 1.000} \\
\bottomrule
\end{tabular}
}
\end{table}

\begin{table}[p]
\centering
\caption{Mean calculation times (sec.) to reach convergence over all batches.}
\label{tab::timeparatuck2}
\scalebox{0.6}{
\setlength{\tabcolsep}{6.25pt}
\begin{tabular}{ccccccccccc}
\toprule
  Decomposition & Optimizer & CIFAR-10 & CIFAR-100 & MNIST & COCO & LFW \\
\midrule
CP & ALS & 5.289 & 	4.584 & 	2.710 & 	5.850 & 	4.085	 \\ 
CP & SGD & 1060.455 & 	1019.432 & 	0.193 & 	2335.060 & 	6657.985 \\
CP & NAG & 280.432 & 	256.196 & 	0.400 & 	1860.660 & 	1.317    \\ 
CP & Adam & 2587.467 & 	2771.068 & 	2062.562 & 	6667.673 & 	6397.708 \\ 
CP & RMSProp & 2013.424 & 	2620.088 & 	2082.481 & 	5588.660 & 	4975.279 \\ 
CP & SAGA & 1141.374 & 	1160.775 & 	0.191 & 	3504.593 & 	3692.471 \\ 
CP & AdaGrad & 1768.562 & 	2324.147 & 	959.408 & 	3729.306 & 	6269.536 \\ 
CP & NCG & 315.132 & 	165.983 & 	4.910 & 	778.279 & 	716.355  \\ 
CP & L-BFGS & 2389.839 & 	2762.555 & 	2326.405 & 	5936.053 & 	5494.634 \\ 
CP & VecHGrad & 200.417 & 	583.117 & 	644.445 & 	1128.358 & 	1866.799 \\[.15cm]

DEDICOM & ALS & 21.280 & 	70.820 & 	14.469 & 	55.783 & 	158.946   \\
DEDICOM & SGD & 1826.214 & 	1751.355 & 	1758.625 & 	1775.100 & 	1145.594  \\
DEDICOM & NAG & 30.847 & 	25.820 & 	240.587 & 	43.003 & 	49.518    \\
DEDICOM & Adam & 2105.825 & 	2128.626 & 	1791.295 & 	2056.036 & 	1992.987  \\
DEDICOM & RMSProp & 1233.237 & 	1129.172 & 	993.429 & 	1140.844 & 	1027.007  \\
DEDICOM & SAGA & 27.859 & 	30.970 & 	64.440 & 	28.319 & 	32.154    \\
DEDICOM & AdaGrad & 196.208 & 	266.057 & 	1856.267 & 	2020.417 & 	2027.370  \\
DEDICOM & NCG & 2524.762 & 	644.067 & 	236.868 & 	1665.704 & 	4219.446  \\
DEDICOM & L-BFGS & 1568.677 & 	1519.808 & 	1209.971 & 	1857.267 & 	1364.027  \\
DEDICOM & VecHGrad & 592.688 & 	918.439 & 	412.623 & 	607.254 & 	854.839   \\[.15cm]
  
PARATUCK2 & ALS & 225.952 & 	209.978 & 	230.392 & 	589.437 & 	625.668   \\
PARATUCK2 & SGD & 1953.609 & 	2625.722 & 	2067.727 & 	3002.172 & 	2745.380  \\ 
PARATUCK2 & NAG & 48.468 & 	48.724 & 	285.679 & 	76.811 & 	72.068    \\ 
PARATUCK2 & Adam & 2628.211 & 	2657.387 & 	2081.996 & 	2719.519 & 	2709.638  \\ 
PARATUCK2 & RMSProp & 1407.752 & 	1156.370 & 	1092.156 & 	1352.057 & 	1042.899  \\ 
PARATUCK2 & SAGA & 74.248 & 	70.952 & 	120.861 & 	71.398 & 	86.682    \\ 
PARATUCK2 & AdaGrad & 2595.478 & 	2626.939 & 	2073.777 & 	292.564 & 	2795.260  \\ 
PARATUCK2 & NCG & 150.196 & 	1390.013 & 	928.071 & 	1586.523 & 	82.701    \\ 
PARATUCK2 & L-BFGS & 2780.658 & 	2656.062 & 	2188.253 & 	3522.249 & 	2822.661  \\
PARATUCK2 & VecHGrad & 885.246 & 	1149.594 & 	1241.425 & 	1075.570 & 	1222.827  \\ 
\bottomrule
\end{tabular}
}
\end{table}

\begin{table}[t]
  \caption{PARATUCK2 experimental mean convergence rate per step.}
  \label{tab::conv-rate}
  \centering
  \scalebox{0.65}{
    \setlength{\tabcolsep}{10pt}
    \begin{tabular}{cccccccccc}
	\toprule
    ALS & SGD & NAG & Adam & RMSProp & SAGA & AdaGrad & NCG & L-BFGS & VecHGrad \\
    \midrule
0.958 & 1.004 & 1.010 & 1.009 & 0.992 & 0.994 & 0.983 & 1.376 & 1.452 & \textbf{1.551}  \\ 
  \bottomrule
  \end{tabular}
  }
\end{table}

\section{Conclusion} \label{sec::ccl}
We introduced VecHGrad, a Vector Hessian Gradient optimization method, to solve accurately linear algebra error minimization problems. VecHGrad uses a strong Wolfe's line search, crucial for fast convergence and accurate resolution, with partial information of the second derivative. We conducted experiments on five real data sets, CIFAR10, CIFAR100, MNIST, COCO and LFW, very popular in machine learning and deep learning. We highlighted that VecHGrad is capable to outperform widely-used gradient-based resolution methods, such as Adam, RMSProp or Adagrad, on three different tensor decompositions, CP, DEDICOM and PARATUCK2, offering different levels of linear algebra complexity. We emphasized our experiments with machine learning optimizers since VecHGrad can be easily applied to solve machine learning error minimization problems. Surprisingly, the runtimes of the gradient-based and the Hessian-based optimization methods were very similar as the runtime per step of the gradient-based methods was slightly faster, but their convergence per step was lower. Therefore, gradient-based optimization methods require more iterations to converge. Furthermore, the accuracy of some of the popular schemes, such as NAG, was fairly poor while requiring a similar runtime than that of the other methods. Future work will concentrate on the influence of the adaptive line search to other machine learning optimizers. We observed that the performance of the algorithms is strongly correlated to the performance of the the adaptive line search optimization. Simultaneously, we will look to reduce the memory cost of the adaptive line search as it has a crucial impact on a GPU implementation. We will finally provide a Python and PyTorch public implementations of our method to answer the need of the machine learning and deep learning communities.
\bibliographystyle{./splncs04}
\bibliography{./zzz-mybibliography}
\end{document}